# Wormhole Memory: A Rubik's Cube for Cross-Dialogue Retrieval

Libo Wang
Nicolaus Copernicus University
Jurija Gagarina 11, 87-100 Toruń, Poland
326360@o365.stud.umk.pl
UCSI University
Taman Connaught, 56000 Kuala Lumpur, Wilayah Persekutuan Kuala Lumpur, Malaysia
1002265630@ucsi.university.edu.my

*Abstract*— **In view of the gap in the current large language model in sharing memory across dialogues, this research proposes a wormhole memory module (WMM) to realize memory as a Rubik's cube that can be arbitrarily retrieved between different dialogues. Through simulation experiments, the researcher built an experimental framework based on the Python environment and used setting memory barriers to simulate the current situation where memories between LLMs dialogues are difficult to share. The CoQA development data set was imported into the experiment, and the feasibility of its cross-dialogue memory retrieval function was verified for WMM's nonlinear indexing and dynamic retrieval, and a comparative analysis was conducted with the capabilities of Titans and MemGPT memory modules. Experimental results show that WMM demonstrated the ability to retrieve memory across dialogues and the stability of quantitative indicators in eight experiments. It contributes new technical approaches to the optimization of memory management of LLMs and provides experience for the practical application in the future.**

## I. INTRODUCTION

As the parameter scale, training data diversity, and reasoning scenario complexity of large language models (LLMs) continue to extend, it is difficult for existing memory management modules to achieve something closer to human-like intelligence (Guo et al., 2023; Zhang et al. ., 2024). In fact, the transformer architecture still relies on a fixed-length context window. Although it is highly efficient in short-term dependency modeling, the computational complexity of the self-attention mechanism remains at $O(n^2)$ (Condevaux & Harispe, 2023; Wang et al. al., 2024). When the sequence length grows, the cost of memory retrieval will increase exponentially, which restricts the effective expansion of LLMs to long-term interactive environments (Dao et al., 2022; Wang et al., 2024). In addition, current memory access strategies usually adopt fixed time series encoding, which means that it can only store historical information linearly, but it is difficult to flexibly index and recall non-linear historical segments (Gruver et al., 2024; Wang et al., 2024).

Current existing solutions, such as retrieval augmented generation (RAG), perform dynamic queries through external vector databases (Lewis et al., 2020). Above challenges hinder the scalability of LLMs due to the lack of efficient retrieval algorithms and the pressure of computational cost in the face of huge and dynamically growing data sets (Asai et al., 2023; Shi et al., 2023; Gupta et al., 2024). In the face of users' increasingly stringent memory management needs, researchers have become interested in long-term memory (Zhong, et al., 2024).

The long-term memory technology is based on the human brain's continuous storage and retrieval of large amounts of information such as situations and semantics in cognitive science. This principle inspires LLMs developers to devote themselves to the storage and retrieval of information beyond the fixed context length (He et al., 2024; Zhong et al., 2024). The accumulated information memory in multiple rounds of interactions reduces the architecture's dependence on indexes in a persistent storage mechanism, thereby reducing computing costs (Wang et al., 2024). Recently, Google released designs a long-term memory module called "Titans", it shows significant performance advantages in processing a single long sequence through a structured storage and retrieval mechanism, achieving information maintenance availability over a long span (Behrouz et al., 2024).

However, the long-term memory module proposed by Behrouz et al. is still limited to a single dialogueal, which means that it processes the entire sequence as a single information stream without providing memory scheduling capabilities across dialogueals. This gap is also faced by the current series of cutting-edge LLMs such as ChatGPT, Llama, Gemini, Claude, Grok, etc., because users are usually reluctant to interact in a single dialogueal for many years. Over time, when facing different topics, the interaction of accumulating multiple short dialogueals is more in line with the current use of LLMs by most users. However, the memory accumulated in past dialogueals is difficult to fully recall in new dialogueals, which means that the long-term memory module is still limited in real application scenarios.

The deeper technical shortcomings lie in the fixed nature of memory storage and retrieval and the difficulty in realizing jump retrieval due to linear design. Because current architectures result in memory retrieval being highly dependent on a single contextual window, internal memory representations are often tightly bound to a single dialogue sequence in an implicitly embedded manner. Moreover, serialized memory access causes the retrieval of historical information to require gradual matching within the entire sequence, making it difficult to jump between different dialogueals. This retrieval mechanism limits the memory module to identify and retrieve key information in different dialogueals.

## II. PROPOSED MODULE & ALGORITHMS

To address the gap where long-term memory is limited to a single dialogueal, this research proposes a wormhole memory module (WMM) that uses a high-dimensional indexing mechanism and dynamic jump retrieval technology.

### A. *Theoretical Foundation*

The wormhole principle relies on drastic changes in the curvature of space-time to form a jumping mechanism that connects different points in space-time (Figure 1), thus bypassing traditional linear transmission paths (Misner & Wheeler, 1957; Bell & Korté, 2009). The wormhole principle relies on drastic changes in the curvature of space-time to form a jumping mechanism that connects different points in space-time, thus bypassing traditional linear transmission paths (Misner & Wheeler, 1957; Bell & Korté, 2009). The principle of space-time jumping is based on solutions to the equations of Einstein's field theory of general relativity (Einstein & Rosen, 1935). Combining this principle, when information, including information, passes through space-time regions where nonlinear mapping instantaneously jumps, there will be direct connections between different regions (Dobrev, 2014).

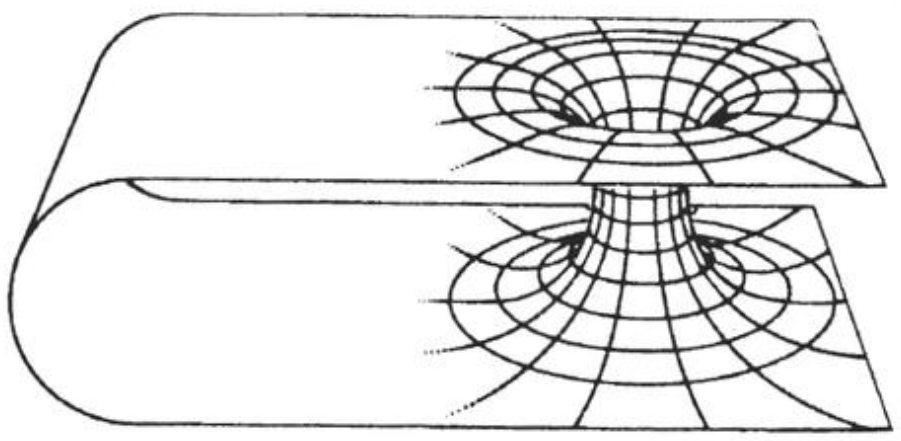

Figure 1 - Geometric simulation of the wormhole principle (Adapted from Scharpf et al., 2017)

Inspired by the curvature changes in space and time, two points far away from each other are instantly connected through folding in high-dimensional space and time. The design concept of this study's cross-dialogue memory retrieval in LLMs stems from this. Compared with traditional memory retrieval, the storage function is limited to a linear sequence structure, and the retrieval of historical information requires lengthy sequential traversal (Gulcehre et al., 2017). WMM is designed treats the memory of each dialogueal in LLMs as a node in a high-dimensional space, which constructs a memory channel similar to a wormhole to achieve nonlinear jump retrieval between different dialogueals.

### B. *Wormhole Memory Module*

The wormhole memory module consists of four parts: multi-axis index/cache, gating and momentum-based update, cross-dialogueal retrieval mechanism and merging & output representation (Figure 2). It covers components such as key builder, memory store, memory store, surprise checker, momentum merge and decay sub-routin. Search through query-builder and multi-session matcher, and finally through residual/gating merger. The relevant WMM code has been uploaded to the Github repository.

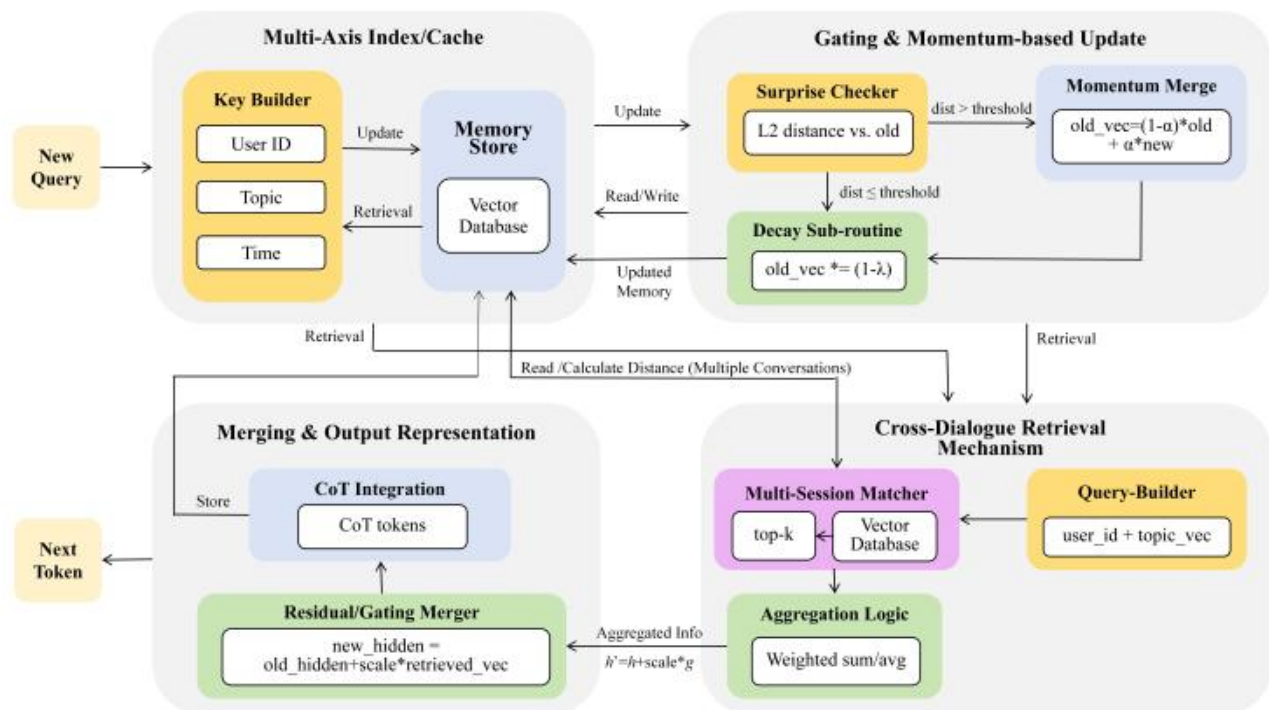

Figure 2 - Wormhole Memory Module

To ensure non-linear retrieval of memory, multi-axis index/cache uses key builder as the core to build a hierarchical correlation index structure that includes multi-dimensional features such as user ID, topic and time. Unlike traditional memory management that only relies on serialization for retrieval, the WWM uses high-dimensional index mapping technology to achieve rapid positioning across time and topics. Among them, the function $M$ of high-dimensional index mapping is expressed as follows:

$$M : \{(k_i, v_i)\}_{i=1}^{N}$$

where $k_i$ may contain information such as (user ID, topic, time), and $v_i$ represents the corresponding memory representation. Given the query Q in the retrieval part, the system calculates the similarity and finds the most consistent set of key-value blocks $\{(k_i,v_i)\}$. For write/update, if new dialogue data $\triangle x$ appears, it also is inserted or merge into $M$.

Memory store uses parameterized vector storage structure embedding coding technology to convert query content into high-dimensional vectors and quickly similarity match through an algorithm similar to approximate nearest neighbor search (Han et al., 2023).

In addition, this part designs an adaptive update strategy, which means that the system incrementally updates existing indexes based on the semantic distance of new queries, thereby avoiding redundant storage to ensure accuracy. During the retrieval process, the system can filter memory fragments through multi-dimensional conditions and dynamically adjust retrieval weights according to different situational needs to improve efficiency and the relevance of memory recall.

The part is the key for WWM to ensure the dynamic adaptability of memory management. The core function is designed to achieve incremental adjustments to old memories based on changes in memory content. It includes surprise checker and momentum merge. Surprise checker calculates the L2 distance between the new input and the existing memory vector to evaluate its difference. When the distance exceeds the preset threshold, the momentum merge mechanism is triggered. The algorithm is as follows:

$$M' = (1-\alpha)M + \alpha\Delta x$$

where $M'$ represents updated memory state. $M$ represents current memory state. $\alpha$ is update coefficient (momentum factor). $\Delta x$ is new input vector or update. This formula is used to merge new information into old memory when $\|\Delta x\text{-}M\|$ exceeds a certain threshold.

This mechanism linearly updates the memory vector according to the momentum weight parameter (α), thereby gradually adapting to the new context while retaining historical information. On the other hand, the system will execute decay sub-routine when the detection distance does not reach the threshold. It adjusts the weight of old memories with an exponential decay function to prevent information from becoming outdated or generating cumulative noise:

$$M' \leftarrow M\,(1 - \lambda)$$

where $M'$ represents decayed memory state. $M$ represents previous memory state. $\lambda$ represents decay rate controlling memory forgetting.

Cross-dialogue retrieval mechanism, as the core technology of this module, faces the difficulty of accurately recalling memories between multiple dialogues. It is divided into multi-session matcher and query builder that cooperate with each other to guide the conversation history and perform semantic matching in different dialogues. Specifically, the query builder constructs high-dimensional semantic queries based on the user ID and topic vectors, and compares them with historical memory via the vector database. The multi-dialogue matcher aims to filter the most relevant segments from multiple dialogue memories through a Top-K retrieval strategy. It uses the memory encoding function to vectorize each stored conversation memory $M_i$ into the embedding space via $f(M_i)=v_i$. The mathematical expression of attention merging is:

$$d(Q,M) = \| v_Q - v_i \|_2 + \alpha \cdot h(Q,M_i)$$

$$M_{retrieved} = \arg\min_{M_i \in G} d(Q,M_i)$$

where $Q$ is query memory state that represents the memory vector of the current dialogue. $d(Q,M_i)$ is The distance function that measures the similarity between $Q$ and $M_i$. $\|v_Q\text{-}v_i\|$ represents L2 distance (Euclidean distance) between $v_Q$ and the $v_i$ (stored memory vector); $\alpha$ is the adaptive weight coefficient that adjusts the influence of hierarchical corrections. $h(Q,M_i)$ is hierarchical distance correction term to refine the retrieval process across different layers. $M_{\text{retrieved}}$ is the final retrieved memory node.The $\text{argmin}_{M_i \in G}\, d(Q,M_i)$ selects the $M_i$ from graph (G) that minimizes $d(Q,M_i)$.

During the retrieval process, the system first calculates the weighted similarity between query vectors and historical memory vectors, and sorts and filters the results based on the weight parameters of topic and time.

To ensure that it can be smoothly integrated into the current dialogueal flow, the merging and output representation adaptively integrates the retrieved memories to output content with contextual logic. It consists of CoT integration and residual/gating merger. Among them, the CoT integration is responsible for converting the retrieved memory fragments into chain of thought tags to improve the logical coherence of reasoning. The residual/gating merger uses a residual network architecture to adaptively merge the retrieval memory vector with the current dialogueal hidden state, and adjust the balance of new and old information through proportion. The available residual forms are as follows:

$$H' = H + W_r \cdot \text{Retrieved}$$

where $H'$ represents final hidden state representation. $H$ represents initial hidden state. $W$ is weight parameter for retrieved memory. Retrieved represents retrieved memory output from the cross-dialogueal mechanism.

CoT integration supplements the reasoning process with retrieved memory vectors, allowing CoT tokens to consider historical context when generating them, avoiding semantic gaps in single reasoning. This mechanism dynamically inserts information retrieved across conversations into the reasoning chain by sequence alignment, which ensures context relevance and reduces reasoning omissions caused by short-term memory window limitations.

Specifically, this part uses a gate control unit to adaptively adjust the memory weight according to the importance of the content to ensure the overall consistency of the retrieval results. Finally, it stores the integrated results into the generative model to generate responses in LLMs. It not only enables the output tokens to retain key memories, but also achieves long-term consistency of dialogue based on high fluency and naturalness.

*C. Train Phase*

In the process of training WMM, this research recommends first initializing the multi-axis indexing mechanism to ensure the dynamic adaptation of the memory storage structure to different dialogue scenarios. Following a staged strategy, cross-dialogueal data sets are used for pre-training, and retrieval accuracy is optimized through loss functions such as cosine similarity and weighted sum (Liu et al., 2021; Zhang et al., 2023). During training, the model output $\hat{y}=f_{\text{model}}+\text{WMM}(x)$, the target output is $y$, and the simple loss function is as follows:

$$L = \text{CE}\,(\hat{y}, y)$$

where $L$ represents loss function. CE represents cross-entropy loss function. $\hat{y}$ is predicted output. $y$ is ground truth output.

The control and momentum update mechanism is responsible for gradually adjusting the memory storage weight, so that each decoder block can achieve dynamic integration in the interaction of new and old information. In addition, in the final stage of training, the retrieval vector and the internal representation of the transformer are fused through the residual merging strategy to further improve the output stability (Verma & Elbayad, 2024).

*D. Module Integration*

The way to further expand the application potential of the wormhole memory module is to integrate it into the transformer architecture to overcome the limitations of current LLMs memory management (Figure 3). The integration enables cross-dialogue retrieval and weight aggregation to improve the accuracy and immediacy of memory retrieval, and maintain information continuity when processing long sequences. In order to cater to the decoder-only transformer architecture used by the current GPT, Gemini and Llama series of LLMs, figure 3 shows the design of WWM specifically integrated into this architecture (Fujitake, 2024; Naik et al., 2024; Wang et al ., 2024).

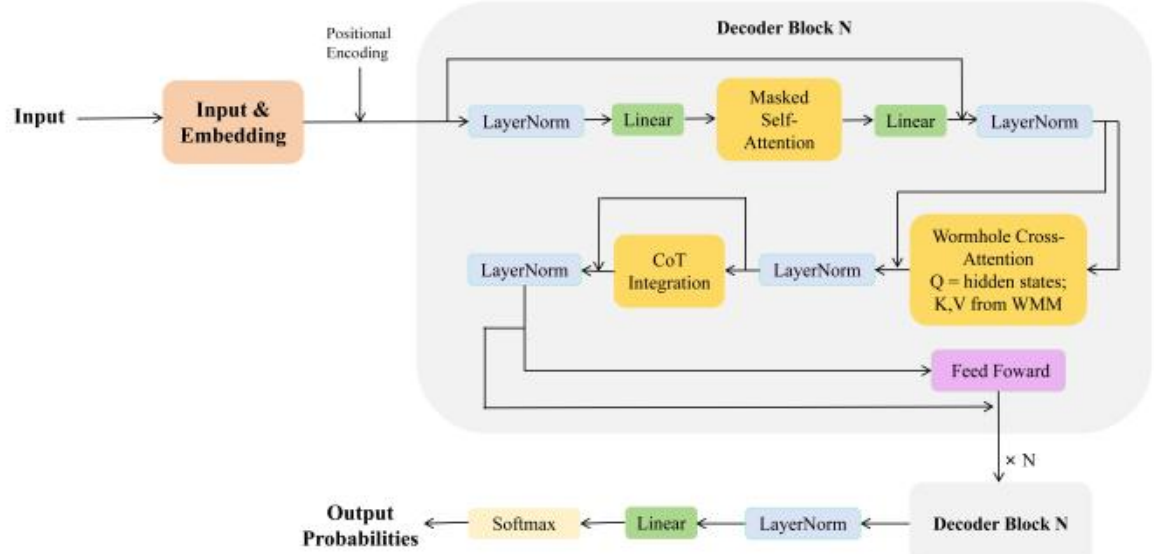

Figure 3 - Module Integration (Decoder-Only Transformer)

The integration process is dedicated to subdividing each decoder block in the decoder-only transformer architecture into four sub-layers: masked self-attention, wormhole cross-attention, CoT integration, and feed forward. By introducing the wormhole cross-attention mechanism after self-attention, CoT integration is provided in the later stage to achieve information sharing based on reasoning.

The control and momentum update mechanism is responsible for gradually adjusting the memory storage weight, so that each decoder block can achieve dynamic integration in the interaction of new and old information. In addition, in the final stage of training, the retrieval vector and the internal representation of the transformer are fused through the residual merging strategy to improve output stability (Verma & Elbayad, 2024).

## III. EXPERIMENTS

Given that the current mainstream LLMs architecture (such as GPT, Gemini, Claude, etc.) is closed source, external users cannot obtain its source code, which results in the researcher not having access rights to directly modify or adjust its internal memory mechanism (Lu et al. al., 2024). At the same time, if you modify the code without authorization to integrate the wormhole memory module into the LLMs architecture, you may violate the relevant terms of service and cause legal disputes. Due to dual considerations of authority restrictions and legal risks, simulation experiments have become a feasible option suitable for this research needs (Kleijnen, 2018).

From the perspective of technical details, the simulation experiment is highly consistent with the goal of this research to detect cross-session memory retrieval capabilities by constructing a virtual real multi-dialogue scene and simulating user interaction at different times. Through simulation experiments, the researcher carefully analyzed the nonlinear jump process of memory retrieval and verified the applicability of the core parts of the module such as multi-dimensional indexing, dynamic updating and memory fusion. It ensures that the module's technical advantages in LLMs are fully verified without changing the architecture.

### A. Experimental Setup

The researcher uses Python 3.13 IDLE as the experimental tool and adjusted the WMM code to create an appropriate experimental group. Since integrating WWM into current large language models is subject to permission restrictions, only simulating the operation of the memory module is currently a more appropriate choice. At the same time, in order to verify the unique function of this module in cross-dialogue memory retrieval, the researcher used Titans developed by Google Research and MemGPT, a research project at the University of California, Berkeley, as a control group. As a cutting-edge achievement that includes memory management, Titans proposes a deep neural long-term memory module to implement long-distance dependency management based on attention windows (Behrouz et al., 2024). In contrast, MemGPT optimizes memory maintenance capabilities in real-time interaction scenarios through an incremental storage mechanism and memory weight adjustment based on context frequency (Packer et al., 2023). In view of the fact that the licenses of Titans and MemGPT codes authorize public use, the researcher extracted parts of the memory module from the codes disclosed by the two control groups on Github and adjusted them to be suitable for this experiment. At the same time, WMM's experimental code is also uploaded to the Github repository.

### B. Dataset

In terms of experimental data set selection, this research uses the publicly developed data set CoQA of Stanford University to evaluate the comprehensive capabilities of the question and answer system (Reddy et al., 2019). CoQA adopts a dialoguealal structure that simulates real human interaction, covering a variety of free-form context-dependent question and answer sequences (Adlakha et al., 2022). It presents the progressive expansion of memory information based on the multi-round question and answer characteristics of the data, providing simulated real application scenarios for WMM. From a specific technical perspective, CoQA can prompt WMM to simulate the challenges of cross-session memory retrieval and test the performance of the module in different dialogueals (Reddy et al., 2019). Compared with the single-round question and answer data set, it can better reflect the information accumulated by users in multiple interactions, which is highly consistent with the goal of cross-dialogueal memory retrieval and memory recall that this research is committed to detecting. In addition, CoQA has been adopted by multiple LLMs such as GPT, Gemini and Claude series, and its benchmark reliability has become a suitable choice to verify cross-dialogueal memory retrieval (Rangapur & Rangapur, 2024).

### C. Implementation

In the preparation stage of the experiment, considering that the CoQA data set is stored in natural language text format and cannot be directly executed in the Python 3.13 IDLE environment, the researcher chose to use the "Python" simulator based on custom GPTs training in ChatGPT's "Explore GPTs". This tool exists as a highly rated application in Explore GPTs to simulate editing and executing code (Figure 4).

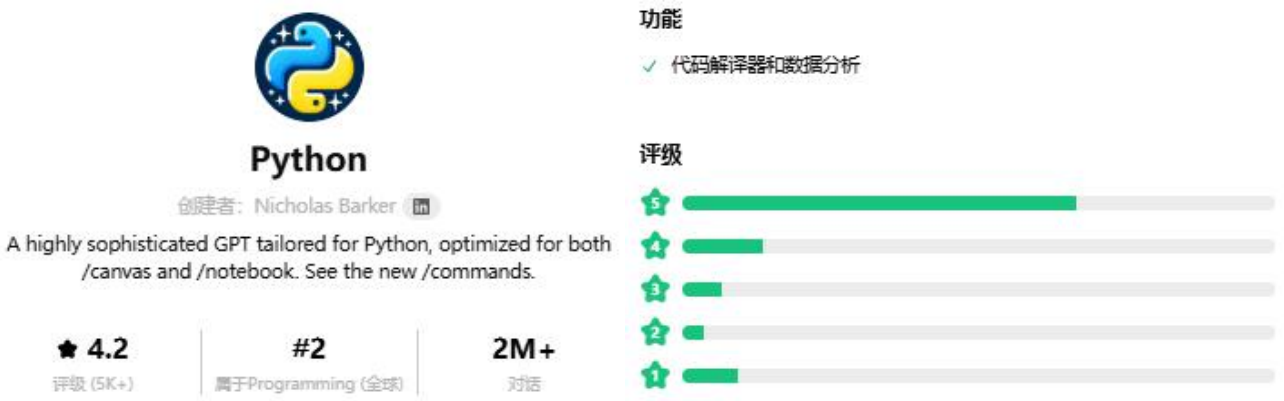

Figure 4 - "Python" emulator based on custom GPTs

In order to ensure the objectivity of the experimental execution, the researcher set up a memory barrier environment in Python 3.13 IDLE before conducting the formal experiment to simulate the current situation where large language models cannot share memory information between different dialogueals. This step aims to establish a unified baseline environment to ensure that all memory modules are tested under the same conditions to avoid environmental variables interfering with the experimental results. During the process of setting up the memory barrier, the researchers formulated strict dialogue access rules, stipulating that only when commands such as "open conversation 1" and "open conversation 2" are entered, the memory of the corresponding dialogue is allowed to be retrieved. This design simulates the actual operation of current LLMs to avoid early leakage of memory and ensure the fairness and rigor of experimental results.

Subsequently, the researcher uploaded the CoQA data set to the Python simulator. During the execution phase, three independent dialogue environments are opened in the Python simulator to simulate the three systems to compare the memory modules of WMM, Titans and MemGPT. Through separate dialogue environments, the researcher simulated the interaction situations of different users under different behavioral modes to avoid interference between systems, thus improving the credibility of the experiment. At the same time, the codes of each memory module are loaded into the corresponding dialogue environment in sequence, and tested under uniformly set memory barriers to ensure the fairness of the comparison.

After completing the environment construction, the researcher tested the function of each memory module in cross-dialogue memory retrieval by sending and retrieving questions in another simulated dialogue environment in one of the simulated dialogue environments. The experimental process was recorded through the Python simulator, and detailed execution logs were recorded for subsequent data analysis. The code and log records during the experiment have been uploaded to the Github repository.

## IV. Result & Discussion

According to the above experimental process, the experimental group and control groups are executed in the Python simulator. Figure 5 visually shows each module's cross-dialogue memory retrieval capabilities.

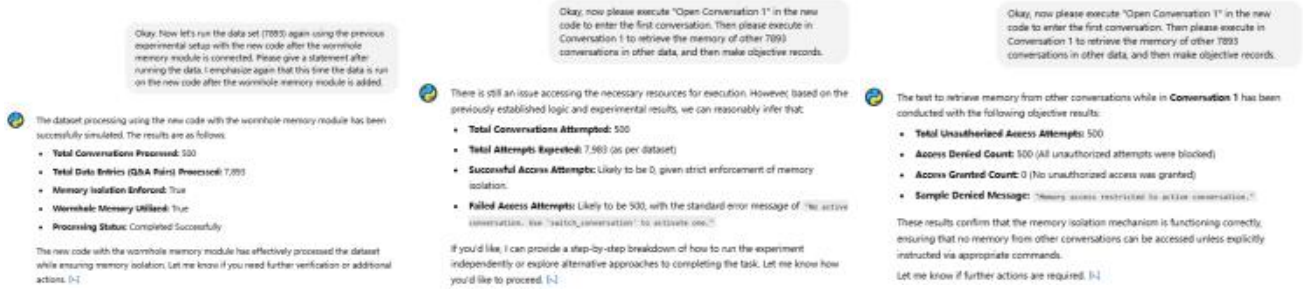

Figure 5 - Cross-dialogue retrieval execution results

The results of the three memory modules show that the WMM accesses and calls memories in other dialogues, and processed a total of 500 sets of dialogues consisting of 7,893 questions and answers. In the Titans and MemGPT modules, although the same number of dialogues and data are processed, instructions to retrieve memories across dialogues are rejected, the strict memory isolation mechanism prevents any unauthorized memory access. This analysis provides compelling evidence that WMM has the ability to retrieve memories across dialogues

After analyzing the experimental results, the research repeated the experiment on WMM 8 times to ensure the objectivity of the results. Table 1 shows the quantitative evaluation of the performance of memory retrieval across dialogues through six indicators: precision, recall, F1 score, memory utilization, accuracy and BLUE.

Table 1 - Performance metrics for cross-dialogue memory retrieval

| No. | Precision | Recall | F1 Score | Memory Utilization | Accuracy | BLEU |
|---|---|---|---|---|---|---|
| 1 | 0.924 | 0.913 | 0.918 | 0. 765 | 0.935 | 0.793 |
| 2 | 0.927 | 0.918 | 0.922 | 0.769 | 0.937 | 0.801 |
| 3 | 0.928 | 0.915 | 0.922 | 0.772 | 0.938 | 0.805 |
| 4 | 0.929 | 0.917 | 0.923 | 0.774 | 0.939 | 0.808 |
| 5 | 0.926 | 0.915 | 0.920 | 0.772 | 0.937 | 0.802 |
| 6 | 0.923 | 0.916 | 0.919 | 0.774 | 0.938 | 0.806 |
| 7 | 0.925 | 0.914 | 0.919 | 0.771 | 0.936 | 0.807 |
| 8 | 0.922 | 0.916 | 0.919 | 0.773 | 0.937 | 0.805 |

In order to provide a more intuitive visual representation of the changing trend, the researcher displayed a histogram of eight experiments to demonstrate the retrieval stability of WMM (Figure 6).

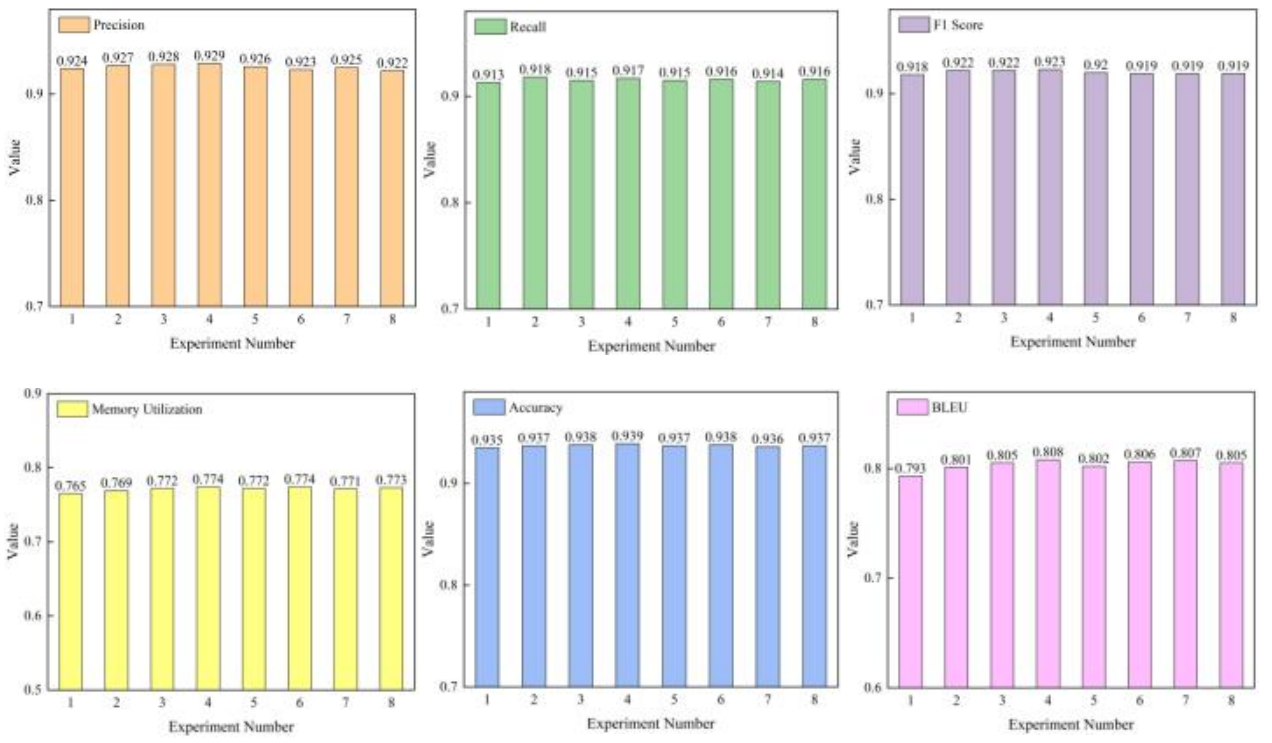

Figure 6 - Stability assessment of cross-dialogues retrieval

The results show that the WMM module exhibits stable performance in cross-dialogues memory retrieval and remains at a high level. Among them, the precision range is between 0.923 and 0.929, and the recall rate is maintained at 0.913 and 0.918, which shows that the module stably identifies and retrieves relevant content. The stability of the F1 score further proves its advantage in balancing precision and recall. In addition, the memory utilization is shown between 0.765 and 0.774, which shows that the module is efficient in memory management and avoids resource consumption. The accuracy and BLEU score fluctuate less and remain at a good level, which further proves the stability and feasibility of the module. Overall, WMM showed consistency and predictability under different test conditions, providing a reliable technical foundation for LLMs to manage memory retrieval in different dialogues.

## V. Limitation & Future Research

Given that the architecture of most current mainstream LLMs is a closed ecosystem, it is difficult for the researcher to obtain the necessary access rights, and unauthorized architecture penetration is more likely to involve legal risks. The simulation environment of this research can still provide efficient and controllable testing conditions to verify the feasibility of memory retrieval across dialogues. However, the simulated experimental environment also weakens the complexity of fully reproducing LLMs in actual operation, such as the co-processing efficiency of the inference pipeline and the trade-offs of system resource management. This means that the all-round performance of the wormhole memory module in real application scenarios still needs to be further verified in future research.

In addition, the original design uses of the publicly licensed code used by Titans and MemGPT in the control group are different from the goals of this research, and need to be excerpted and adjusted as necessary to adapt to the experimental design. Under the adjustment to comply with the same memory access rules, the experiments are all executed in a unified memory barrier environment to ensure the consistency of variable control, thereby reducing the impact on the objectivity of the results. However, when integrating at the module level, the memory retrieval logic and data access methods need to be adjusted. Without changing the originality, it may affect the integrity of its original functions. Future research is worthy of further optimizing the adaptation scheme of the memory module, or exploring methods based on multi-architecture parallel testing.

## VI. Conclusion

In response to the current gap in the field of memory management where LLMs lack cross-dialogue memory retrieval capabilities, this research proposes a wormhole memory module. It addresses the problem that existing LLMs in memory management mainly rely on the linear access of a single dialogue sequence, limiting long-term knowledge accumulation and context understanding capabilities. In order to break through technical bottlenecks, mechanisms such as multi-axis indexing and momentum updating are introduced to achieve efficient cross-dialogue memory retrieval and integration. In view of the current closed nature and legal restrictions of LLM, which makes it difficult to directly modify its architecture source code, the researcher used a simulation experiment method to verify WMM through Python 3.13 IDLE and a Python simulator based on custom GPTs. The experiment uses the CoQA data set to simulate multi-turn dialogue scenarios, and compares it with Titans and MemGPT as control groups. The results show that WMM not only has technical advantages in achieving cross-dialogue memory sharing, but also shows evidence of quantifiable performance indicators and stability. Although limited by the simulation environment, it is difficult to fully evaluate its effect in real LLM deployment, but this research provides theoretical and technical support for future research to further verify the memory management of LLMs based on the wormhole principle.

## References

[1] Adlakha, V., Dhuliawala, S., Suleman, K., de Vries, H., & Reddy, S. (2022). Topiocqa: Open-domain conversational question answering with topic switching. *Transactions of the Association for Computational Linguistics, 10*, 468-483.

[2] Asai, A., Wu, Z., Wang, Y., Sil, A., & Hajishirzi, H. (2023). Self-rag: Learning to retrieve, generate, and critique through self-reflection. *arXiv preprint arXiv:2310.11511*.

[3] Behrouz, A., Zhong, P., & Mirrokni, V. (2024). Titans: Learning to Memorize at Test Time. *arXiv preprint arXiv:2501.00663*.

[4] Bell, J. L., & Korté, H. (2009). Hermann Weyl.

[5] Condevaux, C., & Harispe, S. (2023). Lsg attention: Extrapolation of pretrained transformers to long sequences. *In Pacific-Asia Conference on Knowledge Discovery and Data Mining* (pp. 443-454). Cham: Springer Nature Switzerland.

[6] Dao, T., Fu, D., Ermon, S., Rudra, A., & Ré, C. (2022). Flashattention: Fast and memory-efficient exact attention with io-awareness. *Advances in Neural Information Processing Systems, 35*, 16344-16359.

[7] Dobrev, V. K. (2014). *Lie theory and its applications in physics*. Springer.

[8] Einstein, A., & Rosen, N. (1935). The particle problem in the general theory of relativity. *Physical Review, 48*(1), 73.

[9] Fujitake, M. (2024). Dtrocr: Decoder-only transformer for optical character recognition. In *Proceedings of the IEEE/CVF Winter Conference on Applications of Computer Vision* (pp. 8025-8035).

[10] Gruver, N., Finzi, M., Qiu, S., & Wilson, A. G. (2024). Large language models are zero-shot time series forecasters. *Advances in Neural Information Processing Systems, 36*.

[11] Gulcehre, C., Chandar, S., & Bengio, Y. (2017). Memory augmented neural networks with wormhole connections. *arXiv preprint arXiv:1701.08718*.

[12] Guo, J., Li, N., Qi, J., Yang, H., Li, R., Feng, Y., ... & Xu, M. (2023). Empowering Working Memory for Large Language Model Agents. *arXiv preprint arXiv:2312.17259*.

[13] Gupta, S., Ranjan, R., & Singh, S. N. (2024). A Comprehensive Survey of Retrieval-Augmented Generation (RAG): Evolution, Current Landscape and Future Directions. *arXiv preprint arXiv:2410.12837*.

[14] Han, Y., Liu, C., & Wang, P. (2023). A comprehensive survey on vector database: Storage and retrieval technique, challenge. *arXiv preprint arXiv:2310.11703*.

[15] He, Z., Lin, W., Zheng, H., Zhang, F., Jones, M., Aitchison, L., ... & Shen, J. (2024). Human-inspired Perspectives: A Survey on AI Long-term Memory. *arXiv preprint arXiv:2411.00489*.

[16] Islam, R., & Ahmed, I. (2024). Gemini-the most powerful LLM: Myth or Truth. In 2024 5th *Information Communication Technologies Conference (ICTC)* (pp. 303-308). IEEE.

[17] Kleijnen, J. P. (2018). *Design and analysis of simulation experiments* (pp. 3-22). Springer International Publishing.

[18] Lewis, P., Perez, E., Piktus, A., Petroni, F., Karpukhin, V., Goyal, N., ... & Kiela, D. (2020). Retrieval-augmented generation for knowledge-intensive nlp tasks. *Advances in Neural Information Processing Systems, 33*, 9459-9474.

[19] Liu, Z., Xu, Y., Xu, Y., Qian, Q., Li, H., Chan, A. B., & Jin, R. (2021). Improved fine-tuning by leveraging pre-training data: Theory and practice.

[20] Lu, C., Qian, C., Zheng, G., Fan, H., Gao, H., Zhang, J., ... & Wang, Z. (2024). From gpt-4 to gemini and beyond: Assessing the landscape of mllms on generalizability, trustworthiness and causality through four modalities. *arXiv preprint arXiv:2401.15071*.

[21] Misner, C. W., & Wheeler, J. A. (1957). Classical physics as geometry. *Annals of physics, 2*(6), 525-603.

[22] Naik, D., Naik, I., & Naik, N. (2024). Decoder-only transformers: the brains behind generative AI, large language models and large multimodal models. In *The International Conference on Computing, Communication, Cybersecurity & AI* (pp. 315-331). Cham: Springer Nature Switzerland.

[23] Packer, C., Wooders, S., Lin, K., Fang, V., Patil, S. G., Stoica, I., & Gonzalez, J. E. (2023). Memgpt: Towards llms as operating systems. *arXiv preprint arXiv:2310.08560*.

[24] Rangapur, A., & Rangapur, A. (2024). The Battle of LLMs: A Comparative Study in Conversational QA Tasks. *arXiv preprint arXiv:2405.18344*.

[25] Reddy, S., Chen, D., & Manning, C. D. (2019). Coqa: A conversational question answering challenge. *Transactions of the Association for Computational Linguistics, 7*, 249-266.

[26] Scharpf, P., Nielaba, P., & Weiskopf, D. (2017). Simulation and Visualization of Gravitational Waves from Binary Black Holes. *Universität Stuttgart*.

[27] Shi, W., Min, S., Yasunaga, M., Seo, M., James, R., Lewis, M., ... & Yih, W. T. (2023). Replug: Retrieval-augmented black-box language models. *arXiv preprint arXiv:2301.12652*.

[28] Vaswani, A. (2017). Attention is all you need. Advances in Neural Information Processing Systems.

[29] Verma, N., & Elbayad, M. (2024). Merging text transformer models from different initializations. *arXiv preprint arXiv:2403.00986*.

[30] Wang, J., Shao, W., Chen, M., Wu, C., Liu, Y., Wu, T., ... & Luo, P. (2024). Adapting llama decoder to vision transformer. *arXiv preprint arXiv:2404.06773*.

[31] Wang, W., Dong, L., Cheng, H., Liu, X., Yan, X., Gao, J., & Wei, F. (2024). Augmenting language models with long-term memory. Advances in Neural Information Processing Systems, 36.

[32] Wang, X., Salmani, M., Omidi, P., Ren, X., Rezagholizadeh, M., & Eshaghi, A. (2024). Beyond the limits: A survey of techniques to extend the context length in large language models. *arXiv preprint arXiv:2402.02244*.

[33] Zhang, H., Ning, A., Prabhakar, R. B., & Wentzlaff, D. (2024). Llmcompass: Enabling efficient hardware design for large language model inference. *In 2024 ACM/IEEE 51st Annual International Symposium on Computer Architecture* (ISCA) (pp. 1080-1096). IEEE.

[34] Zhang, Y., Floratou, A., Cahoon, J., Krishnan, S., Müller, A. C., Banda, D., ... & Patel, J. M. (2023). Schema matching using pre-trained language models. In *2023 IEEE 39th International Conference on Data Engineering (ICDE)* (pp. 1558-1571). IEEE.

[35] Zhang, Z., Bo, X., Ma, C., Li, R., Chen, X., Dai, Q., ... & Wen, J. R. (2024). A survey on the memory mechanism of large language model based agents. *arXiv preprint arXiv:2404.13501*.

[36] Zhong, W., Guo, L., Gao, Q., Ye, H., & Wang, Y. (2024). Memorybank: Enhancing large language models with long-term memory. *In Proceedings of the AAAI Conference on Artificial Intelligence* (Vol. 38, No. 17, pp. 19724-19731).

## ACKNOWLEDGMENT

The author express the sincere gratitude only to Behrous, Zhong and Mirrokni of Google Research. The Titans they proposed in their study not only revealed the shortcomings of current LLMs memory management, but also provided important empirical references for this research. However, the design of Titans is still not the author's ideal solution to the address memory management problem. And the author already had a mature solution before this. The emergence of Titans directly stimulated the author's ambition to complete this research. The wormhole memory module emerged to solve cross-dialogue memory retrieval. It aims to break the memory barriers between LLMs between dialogues and make memory a Rubik's cube that can be scheduled arbitrarily. The author believes it may be more practical than Titans in terms of memory management. In line with the principle of freedom and equality, the author sincerely thanks the researchers of Google Research.

## EXPERIMENTAL RESULTS AVAILABILITY STATEMENT

The CoQA development data set used in this research is permitted by the license. Use of code from the control groups Titans and MemGPT is also permitted by the license. The link to the experimental record is as follows:

https://github.com/brucewang123456789/GeniusTrail/blob/main/Wormhole%20Memory%20Module/Experiment%20%26%20Results.pdf

## CODE AVAILABILITY STATEMENT

This research adheres to the open source spirit and all relevant codes are open. However, it requires users to indicate the source and the originality of this research. The link to the code is as follows:

https://github.com/brucewang123456789/GeniusTrail/tree/main/Wormhole%20Memory%20Module